\documentclass[a4paper,journal]{IEEEtran}
\usepackage{amsmath,amsfonts}
\usepackage[hidelinks]{hyperref}
\usepackage{bookmark}
\usepackage[T1]{fontenc}

\usepackage{algorithmic}
\usepackage{algorithm}
\usepackage{array}
\usepackage[caption=false,font=normalsize,labelfont=sf,textfont=sf]{subfig}
\usepackage{textcomp}
\usepackage{stfloats}
\usepackage{booktabs}
\usepackage{url}
\usepackage{tcolorbox}
\usepackage{listings}
\usepackage{verbatim}
\usepackage{afterpage}
\usepackage{multirow}
\usepackage{graphicx}
\usepackage{wrapfig}
\usepackage[
    backend=biber,
    style=numeric,
    sorting=none
  ]{biblatex}
\usepackage{subcaption}

\begin{document}
\pagenumbering{gobble}

\title{A Comparative Analysis of Static Word Embeddings for Hungarian}

\author{Máté Gedeon
\thanks{Máté Gedeon is with the Faculty of Natural Sciences, Budapest University of Technology and Economics, Budapest, Hungary (email: gedeonm01@gmail.com).}
}

\maketitle
\begin{abstract}
This paper presents a comprehensive analysis of various static word embeddings for the Hungarian language, including traditional models such as Word2Vec, FastText, as well as static embeddings derived from BERT-based models using different extraction methods.

We evaluate these embeddings on both intrinsic and extrinsic tasks to provide a holistic view of their performance. For intrinsic evaluation, we employ a word analogy task, which assesses the embeddings' ability to capture semantic and syntactic relationships. Our results indicate that traditional static embeddings, particularly FastText, excel in this task, achieving high accuracy and mean reciprocal rank (MRR) scores. Among the BERT-based models, the X2Static method for extracting static embeddings demonstrates superior performance compared to decontextualized and aggregate methods, approaching the effectiveness of traditional static embeddings.

For extrinsic evaluation, we utilize a bidirectional LSTM model to perform Named Entity Recognition (NER) and Part-of-Speech (POS) tagging tasks. The results reveal that embeddings derived from dynamic models, especially those extracted using the X2Static method, outperform purely static embeddings. Notably, ELMo embeddings achieve the highest accuracy in both NER and POS tagging tasks, underscoring the benefits of contextualized representations even when used in a static form.

Our findings highlight the continued relevance of static word embeddings in NLP applications and the potential of advanced extraction methods to enhance the utility of BERT-based models. This piece of research contributes to the understanding of embedding performance in the Hungarian language and provides valuable insights for future developments in the field. The training scripts, evaluation codes, restricted vocabulary, and extracted embeddings will be made publicly available to support further research and reproducibility.

\end{abstract}


\section{Introduction}

Teaching machines to understand human language is a crucial step toward developing intelligent systems. While one-hot encoding can be effective for small-scale classification tasks, it becomes impractical when the aim is to represent thousands of words and their variations. This limitation led to the development of word embeddings—dense vector representations of words in a continuous space. These embeddings not only represent similar words with similar vectors but also are able to capture complex semantic relationships between them.

Bengio et al. \cite{Bengio} introduced feedforward neural networks with a single hidden layer for language modeling in 2003. These models were capable of learning distributed word representations but suffered from significant scalability issues, limiting their ability to handle large vocabularies. Collobert and Weston \cite{Collobert} addressed this bottleneck in 2008 by refining the training objective and demonstrating the effectiveness of word embeddings pre-trained on large corpora for downstream tasks. Their neural network architecture also served as an inspiration for many subsequent approaches.

Building on the idea that words can be effectively predicted from their context, several word embedding models emerged, including Word2Vec \cite{Mikolov2013}, GloVe \cite{Pennington}, and FastText \cite{Bojanowski}. These models proved highly effective in various natural language processing (NLP) tasks. However, their primary limitation lies in their static nature: each word is assigned a single representation, making it impossible to capture context-dependent meanings (e.g., polysemy). This shortcoming paved the way for the development of dynamic, contextualized word embeddings, exemplified by Transformer-based models \cite{Vaswani} such as BERT \cite{Devlin}, GPT \cite{Radford}, and T5 \cite{Raffel}, which generate word representations that adapt based on surrounding text.

Despite the success of dynamic models, static word embeddings remain a viable option for numerous applications due to their significantly lower computational requirements compared to context-dependent embeddings. In this paper, we analyze the performance of multiple pre-trained static word embeddings for Hungarian, as well as static embeddings extracted from BERT-based models, on both intrinsic and extrinsic tasks.

The main contributions of this paper are as follows:

\begin{itemize}
    \item A comprehensive analysis of the performance of multiple static word embeddings for Hungarian.
    \item An evaluation of static embeddings extracted from BERT-based models.
    \item Public release of the code used in the study\footnote{\url{https://github.com/gedeonmate/hungarian_static_embeddings}}.
    \item Public release of the embeddings extracted from BERT-based models\footnote{\url{https://huggingface.co/gedeonmate/static_hungarian_bert}}.
\end{itemize}

The rest of the paper is organized as follows: Section 2 provides an overview of related work, Section 3 presents the embedding models employed, Section 4 introduces the datasets used in the study, Section 5 describes the experiments conducted, and Section 6 summarizes findings and outlines future directions.

\section{Related Work} 
Multiple methods exist for extracting static embeddings from BERT-based models. The simplest approach involves averaging the token embeddings of a word, when inputting only that word to the model (later referred to as \textit{decontextualized} method). Although easy, this method presents an input to the model, which is different from the training data, potentially leading to suboptimal embeddings. Therefore, more sophisticated methods have been proposed. In this piece of research, we employed two such methods: the \textit{aggregate} method \cite{Bommasani2020} and the \textit{X2Static} method \cite{Gupta2021}. The \textit{aggregate} method pools embeddings of different contexts the word appears in. This presents the model a more natural input than the \textit{decontextualized} method, but as it uses the embedding of the whole context and not just the word itself, it still can lead to suboptimal results. The \textit{X2Static} method, on the other hand, uses a CBOW-inspired static word-embedding approach, leveraging contextual information from a teacher model to generate static embeddings. This method has been shown to outperform the \textit{aggregate} method in multiple tasks.

A Turkish study \cite{Turkish_embeddings} conducted an in-depth analysis of static word embeddings in the language, which had a substantial impact on the chosen methodology in this study. Their findings were similar, finding the \textit{X2Static} method to be the most effective for extracting static embeddings from BERT-based models.

Several studies have contributed to the evaluation of word embeddings in Hungarian. A 2019 paper \cite{Dobrossy2019} assessed the semantic accuracy of word embeddings through word analogy tasks, revealing a substantial performance drop of 50–75\% compared to English. The authors attributed this decline to the high morphological variation in Hungarian and the less stable semantic representations that result from it.

Research on Hungarian contextual embeddings has also gained traction. A comparative study \cite{Ács2021} evaluating huBERT against multilingual BERT models demonstrated that huBERT outperformed its multilingual counterparts in morphological probing, POS tagging, and NER tasks. On the same tasks, but with a focus on the impact of subword pooling, Ács et al. \cite{Ács2021_subword} conducted a comprehensive cross-linguistic analysis of pooling strategies over several languages.

\section{Embedding Models}
\subsection{Word2Vec}
Word2Vec \cite{Mikolov2013} is one of the most influential static word embedding models. It is based on the idea that words appearing in similar contexts have similar meanings. Unlike models focused solely on predicting the next word, Word2Vec considers both preceding and succeeding words within a fixed context window (e.g., five words). The model has two architectures: Continuous Bag of Words (CBOW) and Skip-gram. In CBOW, the goal is to predict a target word from its context, disregarding word order. In Skip-gram, the objective is to predict context words based on a given target word.

Several pre-trained embeddings exist for various languages, including Hungarian. In this study, we used pre-trained embeddings developed for the EFNILEX\footnote{\url{http://corpus.nytud.hu/efnilex-vect/}} project~\cite{Makrai2016}, trained on the combined Hungarian Webcorpus \cite{webcorpus} and the Hungarian National Corpus \cite{hnc} with 600 dimensions.

\subsection{FastText}
In 2018, Bojanowski et al. \cite{Bojanowski} introduced FastText, which provides word embeddings for 157 languages. The model was trained on Wikipedia dumps and the Common Crawl. FastText follows a training procedure similar to Word2Vec but incorporates subword information. The Skip-gram model uses character n-grams, assigning a vector representation to each, and constructs word representations by summing the vectors of the character n-grams present in the word. The full word is included to maintain a unique vector for each word. The CBOW model represents words as bags of character n-grams with position-dependent weights to capture positional information. For this study, we used the 300-dimensional Hungarian embeddings published by FastText.

\subsection{SpaCy}
SpaCy \cite{Spacy} is a Natural Language Processing library implemented in Python and Cython, supporting over 70 languages. It provides pretrained pipelines for tasks such as tagging, parsing, named entity recognition, and text classification. HuSpaCy \cite{HuSpaCy} is the Hungarian adaptation of SpaCy, including pretrained embedding models. We employed the 300-dimensional Hungarian CBOW embeddings\footnote{\url{https://huggingface.co/huspacy/hu_vectors_web_lg}} provided by HuSpaCy, trained on the Hungarian Webcorpus 2.0.

\subsection{ELMo}
ELMo (Embeddings from Language Models) \cite{peters2018} generates contextualized word embeddings using a bidirectional long short-term memory (LSTM) language model. Unlike static embeddings, ELMo captures polysemy and context-dependent meanings, improving performance across multiple NLP tasks. We used the Hungarian ELMo embeddings\footnote{\url{https://github.com/HIT-SCIR/ELMoForManyLangs}} provided by HIT-SCIR, trained on a Hungarian Wikipedia dump and the Hungarian portion of the Common Crawl.

\subsection{BERT-based models}
BERT (Bidirectional Encoder Representations from Transformers) \cite{Devlin} is a transformer-based model that, like ELMo, generates contextualized word embeddings. It is trained using a masked language model (MLM) objective, where the goal is to predict masked words within a sentence, and a next sentence prediction (NSP) objective, where the aim is to determine whether two sentences are consecutive. As a base model, we used huBERT \cite{Nemeskey2021}, trained on the Hungarian Webcorpus 2.0. To have a baseline, we also included XLM-RoBERTa (later referred to as \textit{XLM-R}) \cite{xlmroberta} with the same strategies.

For extracting static embeddings from BERT-based models, we employed the three methods mentioned in \textit{Related Work}. \textit{Decontextualized}, where only the word is inputted to the model, \textit{Aggregate}, which pools embeddings of different contexts the word appears in, and \textit{X2Static}, which uses a CBOW-inspired static word-embedding approach, leveraging contextual information from a teacher model to generate static embeddings. For both the \textit{Aggregate} and \textit{X2Static} methods, we used the same training data, which was a text file, containing one sentence in a line, collected from Hungarian Wikipedia and the Hungarian Webcorpus\footnote{\url{http://mokk.bme.hu/resources/webcorpus/}}. 

Table \ref{tab:models_specs} provides an overview of the models used in the study, including their dimensions and original vocabulary sizes.

\begin{table}[h]
  \centering
  \renewcommand{\arraystretch}{1.2}
  \begin{tabular}{lcc}
      \toprule
      Model & Dimensions & Original Vocabulary \\
      \midrule
      FastText  & 300  & 2,000,000 \\
      huBERT    & 768  & - \\
      EFNILEX   & 600  & 1,896,111 \\
      HuSpaCy   & 300  & 1,524,582 \\
      XLM-R   & 768  & - \\
      ELMo      & 1024 & - \\
      \bottomrule
  \end{tabular}
  \caption{Model specifications}
  \label{tab:models_specs}
\end{table}

\section{Datasets}
\subsection{Intrinsic Evaluation Datasets}
Intrinsic evaluations assess the quality of embeddings independently, rather than through downstream tasks. They typically measure word similarity or evaluate embeddings on word analogy tasks. 

Since no Hungarian word similarity datasets are available, we used the word analogy dataset developed by Makrai \cite{Makrai2015} for intrinsic evaluation. Inspired by the widely used analogy task introduced by Mikolov et al. \cite{mikolov_analogy}, this dataset consists of structured word analogy questions in the form of \textit{A:B :: C:D}, where the relationship between words A and B is expected to be analogous to the relationship between words C and D.

The dataset encompasses multiple linguistic and semantic relations, including morphological (e.g., singular-plural forms, verb conjugations), semantic (e.g., country-capital pairs), and syntactic analogies (e.g., grammatical roles). 

To make comparisons fair, we restricted the embeddings to the vocabulary of their intersection, arriving at a vocabulary of 256 808 words. Table \ref{tab:restrict} provides a comparison of number of questions by category as they were originally, opposed to after the restrictions. There were three categories, where the ratio of the restricted and original questions fell below 90\%: capital-world, family, and gram5-present-participle. For example, the capital-world category contained all the country-capital pairs. As smaller countries capital were not present in the vocabulary, the number of questions in this category significantly reduced. 

\begin{table}[h]
  \centering
  \renewcommand{\arraystretch}{1.2}
  \begin{tabular}{lccc}
      \toprule
      Category & Original & Restricted & Ratio (\%) \\
      \midrule
      capital-common-countries & 190  & 190  & 100.0\% \\
      capital-world            & 13695 & 5995 & 43.8\%  \\
      county-ceter             & 171  & 171  & 100.0\% \\
      currency                 & 435  & 406  & 93.3\%  \\
      family                   & 190  & 136  & 71.6\%  \\
      gram1-adjective-to-adverb & 780  & 780  & 100.0\% \\
      gram2-opposite           & 435  & 435  & 100.0\% \\
      gram3-comparative        & 780  & 780  & 100.0\% \\
      gram4-superlative        & 780  & 780  & 100.0\% \\
      gram5-present-participle & 780  & 496  & 63.6\%  \\
      gram6-nationality-adjective & 820 & 741 & 90.4\% \\
      gram7-past-tense         & 780  & 780  & 100.0\% \\
      gram8-plural-noun        & 780  & 780  & 100.0\% \\
      gram9-plural-verb        & 780  & 780  & 100.0\% \\
      \bottomrule
  \end{tabular}
  \caption{Comparison of Original and Restricted Number of Questions by Categories}
  \label{tab:restrict}
\end{table}

\subsection{Extrinsic Evaluation Datasets}
For extrinsic evaluation, we used NYTK-NerKor \cite{nerkor}, a Hungarian gold-standard named entity-annotated corpus containing 1 million tokens. This dataset also includes POS tags, making it suitable for both Named Entity Recognition (NER) and Part-of-Speech (POS) tagging tasks. The corpus is divided into training, development, and test sets using an approximately 80-10-10 split. It comprises texts from diverse genres, including fiction, legal documents, news, web sources, and Wikipedia. 

The fiction subcorpus includes novels from the Hungarian Electronic Library (MEK) and Project Gutenberg, as well as subtitles from OpenSubtitles. Legal texts originate from sources such as the EU Constitution, documents from the European Economic and Social Committee, DGT-Acquis, and JRC-Acquis. The news subcorpus draws from the European Commission's Press Release Database, Global Voices, and the NewsCrawl Corpus. Web texts are sourced from the Hungarian Webcorpus 2.0.

Named entity annotations are divided into four categories: Person, Location, Organization, and Miscellaneous. Although an updated version of the corpus \cite{nerkor_new} introduces approximately 30 entity types, we used the original version, as these four categories are most commonly utilized in NER tasks.

\section{Experiments}

\subsection{Intrinsic Evaluation}
To assess performance on the analogy task, we employed two evaluation metrics. The first metric was overall accuracy, defined as the proportion of correctly answered questions, where the most likely prediction aligns with the correct answer. The second metric was the Mean Reciprocal Rank (MRR), a widely adopted evaluation measure that rewards not only the best prediction but also cases where the correct answer appears among the top-ranked candidates. Given a set of queries, MRR is computed as the average of the reciprocal ranks of the first relevant result for each query:

\[
\text{MRR} = \frac{1}{|Q|} \sum_{i=1}^{|Q|} \frac{1}{\text{rank}_i}
\]

where \(|Q|\) denotes the total number of queries, and \(\text{rank}_i\) represents the rank position of the first relevant document for the \(i\)th query. To ensure computational feasibility, we restricted analysis to the top 10 ranked answers per query, assigning a score of 0 when the correct answer was absent. Given that different categories contained varying numbers of questions, we also report the average MRR across categories along with the overall MRR (which is the weighted sum).

For static embedding extraction from BERT-based models, we utilized three previously described methods: \textit{decontextualized}, \textit{aggregate}, and \textit{X2Static}. These are indicated by subscripts in the model names. For instance, huBERT\textsubscript{de} refers to static embeddings extracted using the \textit{decontextualized} method, while huBERT\textsubscript{agg} and huBERT\textsubscript{x2} correspond to the \textit{aggregate} and \textit{X2Static} methods, respectively.

\begin{table*}[h]
  \centering
  \resizebox{0.95\textwidth}{!}{ 
  \begin{tabular}{lcccccccccc}
      \specialrule{.125em}{0em}{.4em}
      Category & FastText & huBERT\textsubscript{x2} & EFNILEX & HuSpaCy & XLM-R\textsubscript{x2} & huBERT\textsubscript{agg}& XLM-R\textsubscript{agg} & ELMo & huBERT\textsubscript{de} & XLM-R\textsubscript{de} \\
      \midrule
      capital-common-countries & \textbf{0.77} & 0.58 & 0.45 & 0.44 & 0.40 & 0.25 & 0.26 & 0.09 & 0.17 & 0.01 \\
      capital-world & \textbf{0.83} & 0.50 & 0.28 & 0.25 & 0.30 & 0.23 & 0.17 & 0.03 & 0.08 & 0.00 \\
      county-center & \textbf{0.88} & 0.76 & 0.31 & 0.47 & 0.25 & 0.18 & 0.07 & 0.09 & 0.24 & 0.00 \\
      currency & \textbf{0.31} & 0.10 & 0.19 & 0.15 & 0.09 & 0.07 & 0.09 & 0.12 & 0.07 & 0.00 \\
      family & 0.66 & \textbf{0.67} & 0.40 & 0.59 & 0.46 & 0.22 & 0.25 & 0.33 & 0.30 & 0.05 \\
      gram1-adjective-to-adverb & 0.63 & 0.59 & 0.37 & 0.61 & \textbf{0.78} & 0.20 & 0.26 & 0.13 & 0.25 & 0.07 \\
      gram2-opposite & \textbf{0.43} & 0.16 & 0.29 & 0.24 & 0.17 & 0.01 & 0.04 & 0.10 & 0.07 & 0.01 \\
      gram3-comparative & 0.76 & 0.81 & 0.75 & 0.74 & \textbf{0.82} & 0.30 & 0.39 & 0.47 & 0.38 & 0.10 \\
      gram4-superlative & 0.68 & 0.63 & \textbf{0.72} & 0.59 & 0.29 & 0.21 & 0.19 & 0.27 & 0.22 & 0.00 \\
      gram5-present-participle & 0.55 & 0.17 & 0.11 & \textbf{0.72} & 0.12 & 0.02 & 0.02 & 0.01 & 0.15 & 0.00 \\
      gram6-nationality-adjective & \textbf{0.91} & 0.87 & 0.68 & 0.61 & 0.64 & 0.37 & 0.37 & 0.09 & 0.22 & 0.01 \\
      gram7-past-tense & 0.82 & \textbf{0.95} & \textbf{0.95} & 0.86 & 0.91 & 0.18 & 0.17 & 0.82 & 0.49 & 0.01 \\
      gram8-plural-noun & \textbf{0.77} & 0.61 & 0.66 & 0.68 & 0.66 & 0.28 & 0.34 & 0.67 & 0.70 & 0.06 \\
      gram9-plural-verb & 0.94 & \textbf{0.97} & 0.95 & 0.87 & 0.87 & 0.67 & 0.68 & 0.86 & 0.55 & 0.02 \\
      \midrule
      Average MRR & \textbf{0.71} & 0.60 & 0.51 & 0.56 & 0.48 & 0.23 & 0.24 &  0.29 & 0.28 & 0.02 \\
      Overall MRR & \textbf{0.77} & 0.58 & 0.46 & 0.46 & 0.45 & 0.24 & 0.23 & 0.22 & 0.22 & 0.02 \\
      Overall accuracy & \textbf{0.71} & 0.49 & 0.39 & 0.38 & 0.37 & 0.18 & 0.18 & 0.18 & 0.17 & 0.01 \\
      \specialrule{.125em}{0em}{.4em}
  \end{tabular}
  }
  \caption{Comparison of Performance Across Models (MRR)}
  \label{tab:model_comparison}
\end{table*}

Table \ref{tab:model_comparison} presents the results of the analogy tasks. FastText exhibited superior performance, achieving an overall accuracy of 71\% and an MRR score of 0.77. The second-best model, huBERT, attained 49\% accuracy and an MRR of 0.58. The remaining models formed two distinct groups based on overall accuracy. The first group, comprising EFNILEX, HuSpaCy, and XLM-R\textsubscript{x2}, yielded an MRR of approximately 0.46 and an accuracy of around 38\%. The second group, including huBERT\textsubscript{agg}, XLM-R\textsubscript{agg}, ELMo, and huBERT\textsubscript{de}, exhibited an MRR around 0.23 with an accuracy near 18\%. XLM-R\textsubscript{de} performed the worst, with both metrics falling below 2\%, indicating that the \textit{decontextualized} embedding extraction was unsuitable for this task.

Analyzing overall accuracy, we observe that among the top four models are the three static embedding models, along with only one BERT-based model. This suggests that static embeddings remain competitive for intrinsic tasks despite their lack of contextual adaptability compared to transformer-based models. Notably, huBERT\textsubscript{x2} performed comparably to static embeddings, indicating that the \textit{X2Static} method is a promising technique for deriving static representations from BERT-based models.

The average MRR scores follow a similar pattern to overall accuracy but reveal a more nuanced ranking, as it considers every category equally important, regardless of its size. While the top five models remain consistent, the performance gaps narrow. This discrepancy is likely due to category imbalances, where some categories contain significantly more questions than others. Among the lower-performing models, ordering shifts: ELMo surpasses both \textit{aggregate} models, while huBERT\textsubscript{de} outperforms huBERT\textsubscript{agg}.

Examining the performance of the three embedding extraction methods in terms of overall accuracy, a clear trend emerges. The \textit{decontextualized} method consistently underperforms in both huBERT and XLM-R. The \textit{aggregate} method offers a marginal improvement over \textit{decontextualized} in huBERT, yet it demonstrates a substantial boost in XLM-R. This improvement may stem from XLM-R's multilingual nature, where isolated Hungarian words may provide insufficient input, whereas contextualized sentences enable better representations. The \textit{X2Static} method outperforms both alternatives across both models, approaching the effectiveness of dedicated static embeddings. However, when considering average MRR scores, while \textit{X2Static} remains the best performer, huBERT\textsubscript{de} surpasses huBERT\textsubscript{agg}.

Category-level results reveal substantial variation in model performance. FastText dominates, achieving the highest scores in 7 out of 14 categories, with the largest margins observed in \textit{capital-common-countries},  \textit{capital-world}, and \textit{gram2-opposite}. HuBERT prevails in three categories (\textit{family}, \textit{gram9-plural-verb}, and \textit{gram7-past-tense}), drawing with EFNILEX in the latter. EFNILEX also leads in \textit{gram4-superlative}. HuSpaCy stands out in \textit{gram5-present-participle}, significantly outperforming other models. Notably, this category experienced the most substantial question reduction (37\%) due to vocabulary constraints. HuSpaCy achieved a 72\% MRR score in this category, despite its average MRR of 0.56.

XLM-R\textsubscript{x2} demonstrates strong performance, leading in the \textit{gram1-adjective-to-adverb} category and slightly outperforming others in \textit{gram3-comparative}. An interesting observation is that while ELMo and huBERT\textsubscript{de} exhibit similar overall performance, their category-level results diverge significantly, with each model excelling in different areas. In total, static embeddings win 10 of 14 categories, BERT-based models obtain three victories, and one category results in a tie. These findings underscore that category-level performance can vary considerably, making overall scores an incomplete representation of a models strengths.

When analyzing embedding extraction techniques for huBERT at the category level, \textit{X2Static} consistently outperforms the other methods in all but one category (\textit{gram8-plural-noun}), where the \textit{decontextualized} method proves to be superior. Comparing \textit{aggregate} and \textit{decontextualized} approaches, results are less conclusive. For huBERT, \textit{aggregate} excels in only four categories, while \textit{decontextualized} leads in nine, with one tie. XLM-R follows a more distinct hierarchy, with \textit{X2Static} outperforming all other methods, \textit{aggregate} ranking second, and \textit{decontextualized} consistently underperforming.

These results collectively highlight the effectiveness of static embeddings for analogy tasks, the viability of the \textit{X2Static} approach for extracting static embeddings from transformer-based models, and the varying impact of embedding extraction techniques across different linguistic categories.

\subsection{Extrinsic Evaluation}

For the extrinsic evaluation, we employed a single-layer bidirectional LSTM with a dropout rate of 0.5. The bidirectional LSTM was chosen for its ability to capture contextual information from both directions, which is particularly beneficial for tasks such as Named Entity Recognition (NER) and Part-of-Speech (POS) tagging. To assess the performance of the embeddings, we experimented with varying hidden sizes (1, 2, 4, 8, 16, 32, and 64). While increasing the hidden size generally enhances model performance, the objective was to evaluate the embeddings themselves, including their behavior under constrained settings.

To handle out-of-vocabulary words in the training data, we represented them using vectors sampled from a normal distribution with a mean of 0 and a standard deviation of 0.6 ($\mathcal{N}(0, 0.6)$). The model architecture concluded with a softmax activation function. Training was conducted for five epochs with a batch size of 32, utilizing the Adam optimizer with its default parameter settings. Categorical cross-entropy was employed as the loss function.

Table \ref{tab:ner_comparison} summarizes the performance of the NER models on the test set. As the primary objective was to assess the quality of the embeddings rather than to achieve state-of-the-art results, accuracy was selected as the evaluation metric. While suitable for the purposes of our comparative analysis, it should be noted that this metric is not directly comparable to those reported by leading models in the field, which are typically evaluated using the F1 score \cite{korpusz2022}.

\begin{table}[ht]
  \centering
  \resizebox{0.48\textwidth}{!}{
  \begin{tabular}{lccccccc}
      \specialrule{.125em}{0em}{.4em}
      \multirow{2}{*}{Model Name} & \multicolumn{7}{c}{Hidden Size} \\
      \cmidrule(lr){2-8}
      & 1 & 2 & 4 & 8 & 16 & 32 & 64 \\
      \midrule
      ELMo & 94.81 & \textbf{96.11} & \textbf{96.44} & \textbf{97.01} & \textbf{97.39} & \textbf{97.54} & \textbf{97.62} \\
      huBERT\textsubscript{x2} & 93.55 & 95.42 & 96.02 & 96.68 & 97.24 & 97.40 & 97.49 \\
      RoBERTa\textsubscript{x2} & 94.79 & 95.48 & 96.16 & 96.55 & 97.20 & 97.33 & 97.44 \\
      huBERT\textsubscript{de} & \textbf{94.89} & 94.12 & 96.22 & 96.76 & 97.16 & 97.24 & 97.38 \\
      HuSpaCy & 94.16 & 94.71 & 95.84 & 96.23 & 96.54 & 96.92 & 97.13 \\
      huBERT\textsubscript{agg} & 94.92 & 94.97 & 95.16 & 95.95 & 96.61 & 96.94 & 97.01 \\
      EFNILEX & 94.48 & 94.86 & 95.49 & 96.09 & 96.46 & 96.71 & 96.80 \\
      FastText & 94.59 & 94.89 & 95.62 & 96.32 & 96.56 & 96.68 & 96.78 \\
      XLM-R\textsubscript{agg} & 85.39 & 94.67 & 94.37 & 95.07 & 95.12 & 95.30 & 95.68 \\
      XLM-R\textsubscript{de} & 94.79 & 88.59 & 94.97 & 95.03 & 95.10 & 95.22 & 95.37 \\
      \specialrule{.125em}{.4em}{0em}
  \end{tabular}
  }
  \caption{Model performance across hidden sizes for NER (\%)}
  \label{tab:ner_comparison}
\end{table}

\begin{figure}[ht]
  \centering
  \includegraphics[width=0.9\linewidth]{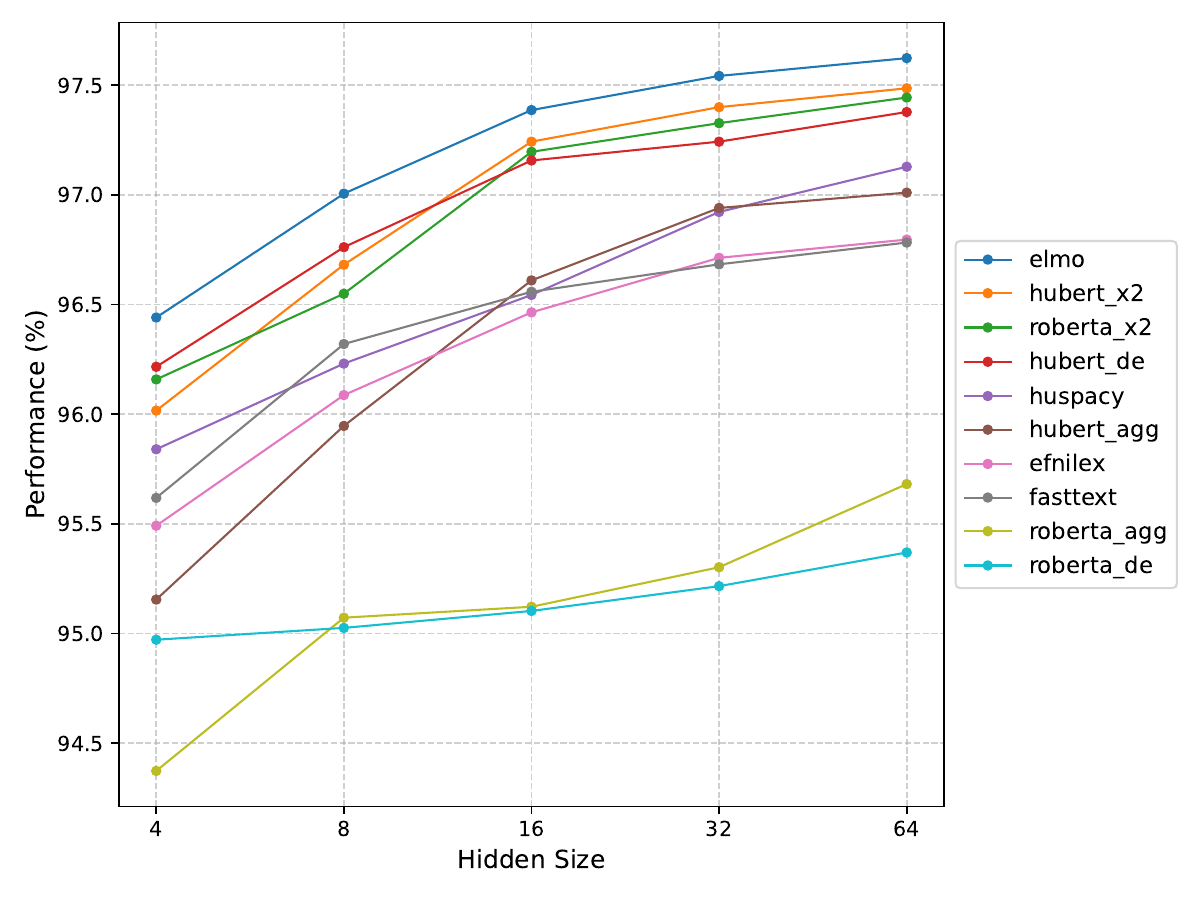}
  \caption{Model performance across hidden sizes for NER (\%)}
  \label{fig:ner_plot}
\end{figure}

As shown in Table~\ref{tab:ner_comparison}, most models demonstrate a consistent improvement in performance with increasing hidden size. However, exceptions to this trend include XLM-R\textsubscript{agg} (between hidden sizes 2 and 4), as well as huBERT\textsubscript{de} and XLM-R\textsubscript{de} (between sizes 1 and 2), with the latter experiencing a particularly sharp decline. To complement the tabular data, Figure~\ref{fig:ner_plot} offers a visual summary that aids intuitive comparison, albeit with a slight trade-off in numerical precision. For clarity, hidden sizes 1 and 2 were excluded from the plot due to their disproportionate effect on the plot scale, which hindered the visibility of differences among the larger hidden sizes.

Regarding overall performance, XLM-R\textsubscript{agg} and XLM-R\textsubscript{de} were the lowest-performing models, indicating their relative inefficacy for this task. In contrast—and somewhat unexpectedly—ELMo emerges as the top-performing model across all hidden sizes of two and above, despite its weaker results in the analogy task.  It may come from the fact, that ELMo is originally trained using a bidirectional LSTM, similarly to the model used in the extrinsic evaluations. For larger hidden sizes, it is followed by the two \textit{X2Static} models, huBERT\textsubscript{x2} and XLM-R\textsubscript{x2}, which maintain close performance levels but exchange rankings depending on the hidden size. Notably, at smaller hidden sizes, huBERT\textsubscript{de} outperforms both \textit{X2Static} models. The remaining four models—HuSpaCy, huBERT\textsubscript{agg}, EFNILEX, and FastText—consistently occupy the middle range, positioned between the top four and the bottom two, with their relative rankings fluctuating based on hidden size.

In terms of extraction methodologies, the \textit{X2Static} approach remains the most effective, consistent with the intrinsic task results. HuBERT performs better in its \textit{decontextualized} configuration compared to its \textit{aggregate} variant in most cases, likely due to the aggregate model incorporating the full contextual embedding rather than isolating word-level information. XLM-R exhibits an opposite trend, albeit with a less pronounced difference than in the analogy task.

These findings suggest that the top-performing four models are all derived from dynamic embeddings, highlighting their advantage for this task. Nevertheless, purely static embeddings are not far behind in performance, demonstrating competitiveness. Notably, huBERT\textsubscript{agg} is the only BERT-based model to be consistently outperformed—depending on the hidden size—by static embedding models.

\begin{figure}[ht]
  \centering
  \includegraphics[width=0.9\linewidth]{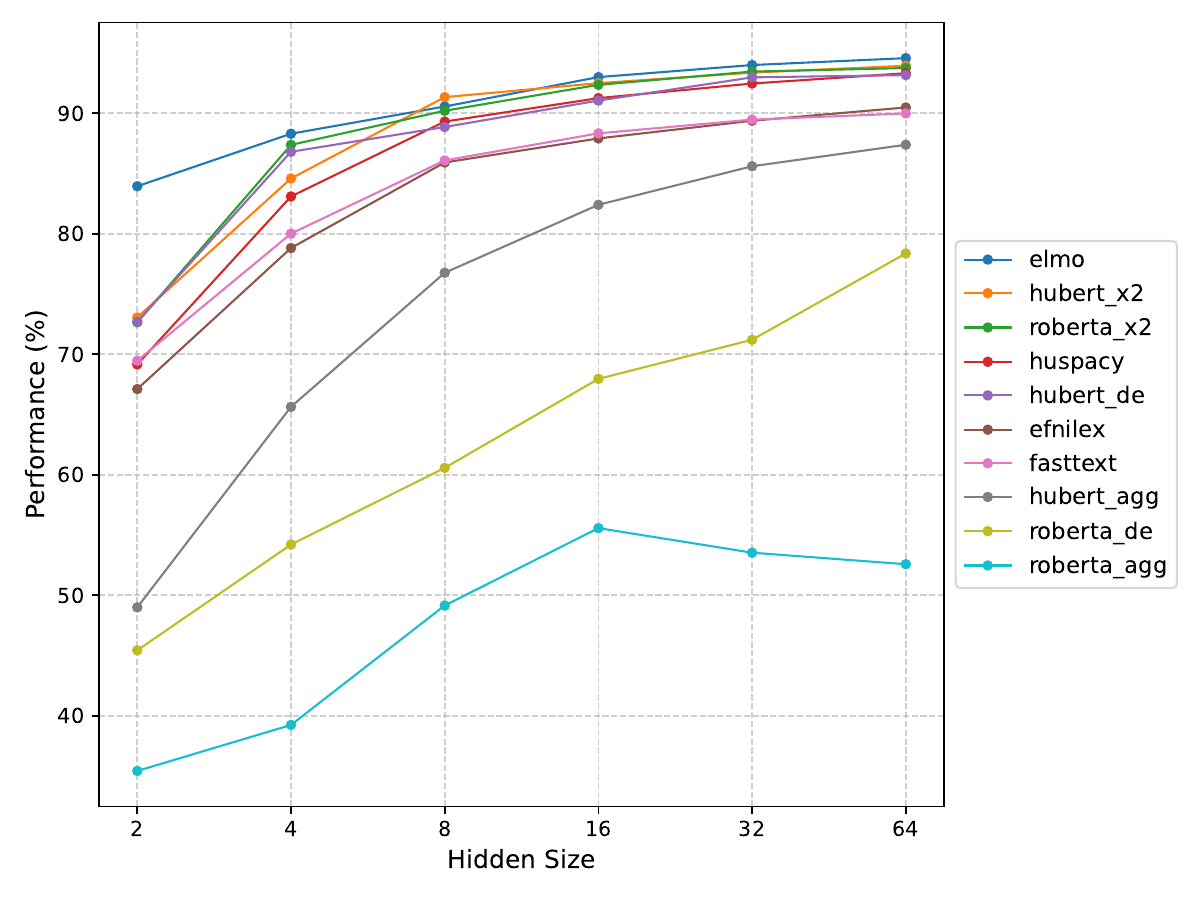}
  \caption{Performance of various models across different hidden sizes in POS tagging (\%)}
  \label{fig:pos_plot}
\end{figure}

Table~\ref{tab:pos_comparison} summarizes the numerical results for the POS tagging task, while Figure~\ref{fig:pos_plot} offers a visual depiction of model performance. As with Figure~\ref{fig:ner_plot}, hidden size 1 is excluded to preserve visual clarity and enable more meaningful comparison across the remaining configurations.

\begin{table}[ht]
  \centering
  \resizebox{0.48\textwidth}{!}{
  \begin{tabular}{lccccccc}
      \specialrule{.125em}{0em}{.4em}
      \multirow{2}{*}{Model Name} & \multicolumn{7}{c}{Hidden Size} \\
      \cmidrule(lr){2-8}
      & 1 & 2 & 4 & 8 & 16 & 32 & 64 \\
      \midrule
      ELMo & 45.69 & \textbf{83.94} & \textbf{88.30} & 90.56 & \textbf{93.00} & \textbf{94.00} & \textbf{94.58} \\
      huBERT\textsubscript{x2} & \textbf{60.40} & 73.06 & 84.59 & \textbf{91.34} & 92.50 & 93.38 & 93.95 \\
      RoBERTa\textsubscript{x2} & 40.89 & 72.65 & 87.38 & 90.21 & 92.36 & 93.46 & 93.76 \\
      HuSpaCy & 51.20 & 69.16 & 83.09 & 89.31 & 91.26 & 92.47 & 93.33 \\
      huBERT\textsubscript{de} & 15.53 & 72.69 & 86.80 & 88.86 & 91.05 & 92.98 & 93.17 \\
      EFNILEX & 38.15 & 67.11 & 78.81 & 85.92 & 87.91 & 89.38 & 90.48 \\
      FastText & 46.09 & 69.46 & 80.01 & 86.09 & 88.33 & 89.47 & 89.99 \\
      huBERT\textsubscript{agg} & 25.87 & 48.99 & 65.64 & 76.76 & 82.40 & 85.60 & 87.39 \\
      XLM-R\textsubscript{de} & 32.19 & 45.43 & 54.20 & 60.57 & 67.95 & 71.20 & 78.37 \\
      XLM-R\textsubscript{agg} & 15.63 & 35.41 & 39.23 & 49.14 & 55.57 & 53.53 & 52.57 \\
      \specialrule{.125em}{.4em}{0em}
  \end{tabular}
  }
  \caption{Performance of various models across different hidden sizes in POS tagging (\%)}
  \label{tab:pos_comparison}
\end{table}

Among the evaluated models, XLM-R\textsubscript{agg} and XLM-R\textsubscript{de} exhibited the lowest performance, while huBERT\textsubscript{agg} outperformed them but remained behind the rest. Consistently, ELMo achieved the highest accuracy across most hidden sizes, followed closely by the two \textit{X2Static} models, huBERT\textsubscript{x2} and XLM-R\textsubscript{x2}.

The remaining models can be grouped into two performance tiers. HuBERT\textsubscript{de} and HuSpaCy formed the stronger pair, demonstrating similar results. EFNILEX and FastText constituted the second tier, with FastText maintaining a slight edge in configurations with less than 64 hidden sizes.

In alignment with findings from other experiments, the \textit{X2Static} method emerged as the best-performing approach. Notably, both huBERT and XLM-R achieved superior results when employing the \textit{decontextualized} method rather than the \textit{aggregate} one.

As observed in the Named Entity Recognition (NER) task, static embeddings derived from dynamic models consistently outperformed purely static models. Despite this trend, the gap in performance remained relatively small, reinforcing the effectiveness of static embeddings in POS tagging tasks.

\section{Conclusion and Future Work}

In this paper, we conducted a comprehensive analysis of various static word embeddings for Hungarian, alongside static embeddings derived from BERT-based models, evaluating them on both intrinsic and extrinsic tasks. To ensure a fair comparison, all models were restricted to a common vocabulary.

For intrinsic evaluation, we employed an analogy task, where FastText demonstrated superior performance, achieving an overall accuracy of 71\% and a Mean Reciprocal Rank (MRR) score of 0.77. Among the BERT-based models, huBERT\textsubscript{x2} emerged as the best performer, with an accuracy of 49\% and an MRR score of 0.58. Notably, the \textit{X2Static} method for extracting static embeddings from BERT-based models outperformed both the \textit{decontextualized} and \textit{aggregate} methods, even rivaling traditional static embeddings in intrinsic evaluations.

For extrinsic evaluation, we utilized a single-layer bidirectional LSTM with varying hidden sizes to assess the effectiveness of the embeddings in downstream tasks. The ELMo embeddings consistently outperformed other models in both Named Entity Recognition (NER) and Part-of-Speech (POS) tagging tasks. The \textit{X2Static} method remained the most effective for extracting static embeddings from BERT-based models, while static embeddings derived from dynamic models outperformed purely static models in both tasks.

This piece of research paves the way for multiple avenues of future exploration. A key direction would be the development of a new intrinsic evaluation dataset for Hungarian, as the analogy question dataset remains the sole benchmark for such evaluations. Additionally, investigating the impact of dimensionality reduction on model performance could yield insights into the trade-offs between efficiency and accuracy, as the models analyzed in this study were trained with varying dimensional settings. Furthermore, the methodology employed here could be extended to evaluate other Hungarian BERT-based models, broadening the scope of comparative analyses. Another significant area of investigation is the exploration of Hungarian GPT-based models (\cite{puli}, \cite{Szentmihalyi2025}), which were not included in this study due to the lack of a clear methodology for extracting static embeddings from such architectures.

In the spirit of open science, the training scripts, evaluation codes, restricted vocabulary, and the extracted huBERT\textsubscript{x2} embeddings are made publicly available, facilitating further research and reproducibility in the field.

\printbibliography
\newpage


\end{document}